# Automatic Time Signature Determination for New Scores Using Lyrics for Latent Rhythmic Structure

Callie C. Liao
*IntelliSky*
McLean, USA
ccliao@intellisky.org

Duoduo Liao
*School of Computing*
George Mason University
Fairfax, USA
dliao2@gmu.edu

Jesse Guessford
*School of Music*
George Mason University
Fairfax, USA
jguessfo@gmu.edu

*Abstract*—There has recently been a sharp increase in interest in Artificial Intelligence-Generated Content (AIGC). Despite this, musical components such as time signatures have not been studied sufficiently to form an algorithmic determination approach for new compositions, especially lyrical songs. This is likely because of the neglect of musical details, which is critical for constructing a robust framework. Specifically, time signatures establish the fundamental rhythmic structure for almost all aspects of a song, including the phrases and notes. In this paper, we propose a novel approach that only uses lyrics as input to automatically generate a fitting time signature for lyrical songs and uncover the latent rhythmic structure utilizing explainable machine learning models. In particular, we devise multiple methods that are associated with discovering lyrical patterns and creating new features that simultaneously contain lyrical, rhythmic, and statistical information. In this approach, the best of our experimental results reveal a 97.6% F1 score and a 0.996 Area Under the Curve (AUC) of the Receiver Operating Characteristic (ROC) score. In conclusion, our research directly generates time signatures from lyrics automatically for new scores utilizing machine learning, which is an innovative idea that approaches an understudied component of musicology and therefore contributes significantly to the future of Artificial Intelligence (AI) music generation.

## I. INTRODUCTION

Music can touch the hearts of any audience, and its transcendental power stems from the fundamental rhythmic structure and melody, the timbre of the instrument, and many more – all of which cooperate in some form of harmony. As a result, music generation has been on the rise along with the increased interest in Artificial Intelligence-Generated Content (AIGC), but to the best of our knowledge, there is still a lack of algorithmic approaches for generating time signatures. In our previous research, we investigated the relations between lyrical keywords and strong beats in songs since strong beats are important to a song's rhythmic structure [1]. This in return motivated us to explore automatic time signature generation for new scores based on lyrical features, as the time signature can be stressed to be one of the most critical rhythmic structural aspects of maintaining a consistent pattern in music, so it is indeed one of the fundamental factors in music generation. However, they may also pose a critical challenge, as researchers behind current music generation methods mostly predetermine time signatures. Overall, time signature determination is vital since it can reveal the latent rhythmic structure that may otherwise be absent from the song. Due to their ability to establish the score's structure, unfitting time signatures will result in less successful music generation and a shift in the entire score, which may form incoherent phrases and result in poorer readability.

Despite that, other methods related to time signatures have been studied. There has been some research on time signature detection and general meter detection, but many require the presence of measures for their algorithms, which would limit the methods to existing scores [2] [3] [4] [5]. This in return would not be applicable to new scores that have yet been assigned a fitting time signature and thus would not have measures. In this paper, we aim to devise reliable time signature generation approaches for new music by using lyrics patterning, natural language processing, and machine learning techniques. More specifically, after processing the data and establishing lyrical patterns and other critical pieces of information, we create various novel features with lyrics, statistical, and musical rhythmic information within the score. Then, we utilize these features as input for machine learning models and proper evaluation metrics to select the most fitting classifier for time signature determination. In the end, only textual data of lyrics is needed as input.

Along the way, time signature determination has created many challenges due to its ambiguity which is especially prevalent in an environment where little to no musical information is present. Time signatures were usually determined through the composer's intuition as well, so developing an algorithmic approach required much analysis regarding the musical structures of songs. Additionally, there was difficulty in selecting the most relevant and useful data in a set of lyrics since keywords, stress patterns, and the consecutive order of these key terms all held valuable information, even if it might not have been considered significant in the past for the literary realm since we have discovered that the literary and musical worlds have similarities but also distinct differences.

Despite these difficulties, we have overcome them with our interdisciplinary backgrounds ranging from music, computer science, artificial intelligence (AI), machine learning, litera-ture, and computational linguistics, all of which are important to AI for music research. With our work that has come to fruition, we present our main contributions as follows:

- Our approach establishes robust and explainable methods for determining time signatures in new scores from only lyrics.
- Our approach unveils the latent rhythmic structure for new scores.
- We developed novel lyrics patterning methods for the lyrical vector pattern of phrases and took into consideration the difference between lyrical and musical characteristics.
- To the best of our knowledge, we are the first to propose a time signature determination approach for new scores using purely lyrics and machine learning.
- Our generation of multiple features encompass lyrical, statistical, and rhythmic information.
- Our methods further reveal that keyword stressed syllables should be emphasized and that they mostly land on strong beats and downbeats.

## II. RELATED WORK

Time signature, an aspect that influences rhythm, is critical to music. In general, [6] states that music changes people's perception of time and discovers that tempo, an aspect of rhythm, influences people's level of arousal. [2] also notes that music meter and tempo are connected and determined together, which indicates that time signature is not only an aspect of rhythm but also influences other rhythmic aspects. Additionally, [7] notes that "time signature determines the number and grouping of beats in a measure" and has a "direct impact on the rhythmic structure of a composition", with some examples being the tension and release of music as well as the different rhythmic patterns that time signature can all affect. This suggests that without a time signature, music may not have a solid rhythmic structure, which reinforces the importance of time signatures, as the absence of them may lead to a decrease in repetition and patterns.

In music generation, a song's time signature is commonly preset by researchers. For [8] [9] [10] [11] [12], their training data is limited to one or a few time signatures such as 2/4 and 4/4, which results in a limited scope for their generated songs. [13] proposes a permutation invariant language model SymphonyNet for symphony music generation, but they utilize predetermined time signatures. It can be inferred that the use of predetermined time signatures is due to the critical role time signature plays in determining a song's rhythmic structure and the lack of research in time signature generation (to the best of our knowledge). Predetermined time signatures may also help simplify the process of music generation so researchers can focus on other aspects, with common focuses being on melodies and lyrics-to-rhythm generation instead.

Despite that, some may not have readily available time signature data, so the time signatures are detected in existing songs or audio recordings before utilizing these songs to generate new melodies. As a result, there is some research on time signature detection methods. In [2], an approach for time signature detection and tempo extraction for Greek traditional music is based on the self-similarity analysis for audio recordings and calculates the numerator and denominator of the time signature separately. Another approach in [3] uses a multi-resolution audio similarity matrix for time signature detection that only depends on the song's musical structure and components such as tempo. It utilizes the repetitive nature of music structure to estimate the number of beats in a measure. Because of the previous paper's success, [4] combines that method with a beat similarity matrix to perform detection on audio recordings. However, all of these are time signature detection methods; hence, they do not consider determining a time signature for a new score.

Similar detection methods have been proposed for meters, which are what time signatures describe as a form of notation for them in a piece or song. [14] proposes that meter arises from groups of rhythmic events that form hierarchical structures consisting of smaller rhythmic and larger metrical units. [5] proposes an approach that helps solve the problem of classifying polyphonic musical audio signals by determining whether the music has a duple or triple meter, thus simplifying the problem. The paper sets the beat as the relevant temporal resolution and focuses on seeking a set of low-level features. The periodicities of those features have the greatest chances of corresponding with specific meters. [15] conducts meter detection using Bach's fugues for automatic music transcription by breaking down a single bar into beats and sub-beats. Although the research presented has certainly contributed to meter detection, the same problem remains for these meter detection methods – they do not consider determining time signatures for completely new scores either.

In [1], the relationships between lyrical keywords and strong beats have been investigated as well as stressed syllables and strong beats in lyrical songs, as strong beats help establish the rhythmic structure of the song that is determined by the time signature. Time signatures also set up the framework for the melody and notes, which are partially dependent on the lyrics for lyrical songs. Thus, since lyrical keywords and phrases are equally integral to time signature determination, our previous work done in [1] serves as a foundation and inspiration for this research, which discovers the importance of keywords and strong beats in AI for music generation. In this paper, time signatures are generated for new scores based on their lyrical content, which to the best of our knowledge vastly differs from previous research in two ways: 1) most only study existing scores, and 2) most time signature determination or meter detection are based on musical and auditory aspects rather than lyrical and textual content. Thus, our approach explores a new facet in the field of AI for music generation.

## III. METHODOLOGIES

Our proposed approach aims to seek the most fitting method for determining time signatures and reveal the latent rhythmic structure of a song, which applies to both new scores and existing scores. In particular, we focus on studying the potential correlations between keywords, stressed syllables, and time signatures because of the frequent association of the keywords with strong beats [1], in which the strong beats are determined through the numerator of the time signature. The architectural framework of this time signature determination approach is illustrated in Figure 1. The framework consists of three sections: data collection and preprocessing, lyrics patterning, and time signature determination. The first section performs lyrical song information extraction as well as web scraping. Then, the second section preprocesses the text and extracts the lyrical keywords, which leads to keyword extraction, phrase patterning, and feature generation. In the third section, features are extracted for machine learning modeling, and imbalanced data is treated. As a result, the classifier with the highest score generated from the evaluation metrics is utilized as the final model.

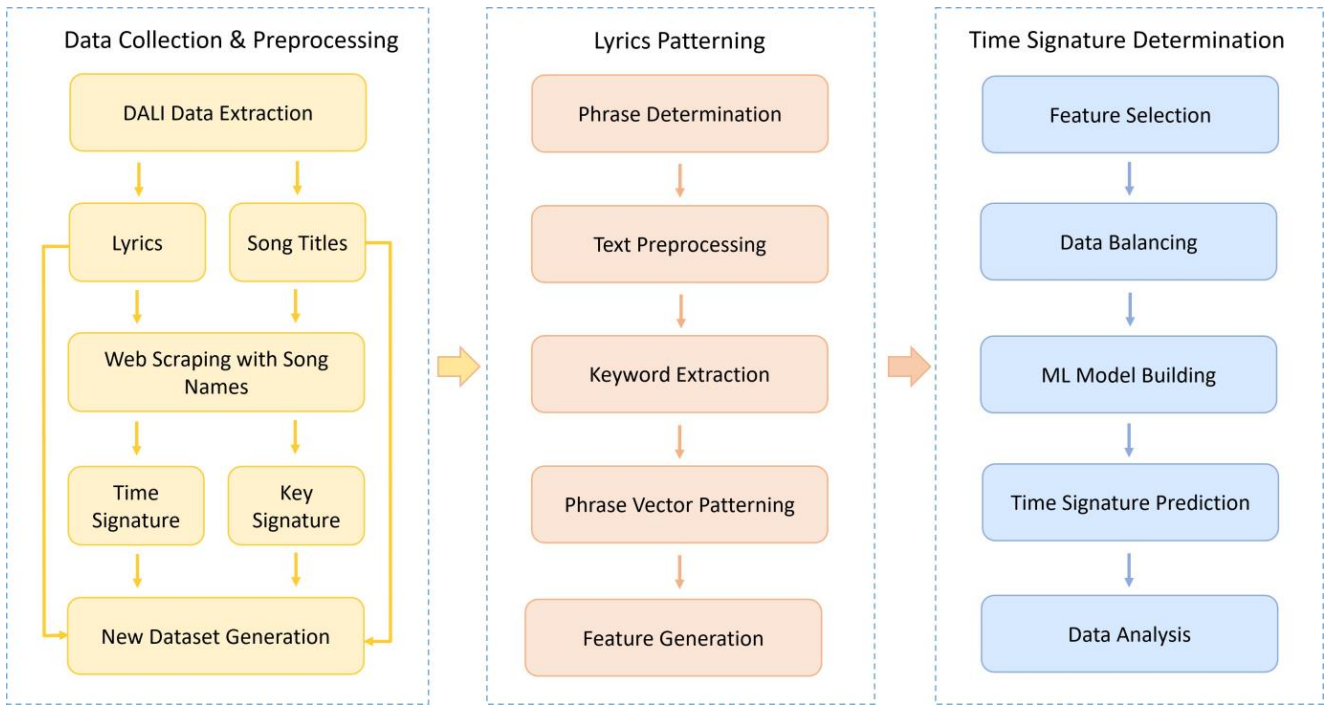

Fig. 1: The framework of time signature determination

### A. Data Collection and Preprocessing

This section extracts the lyrics and song titles from the DALI dataset [16], which does not include time signatures. As a result, data scraping across the Internet is performed for time signature retrieval. The DALI dataset [17] is a multimodal dataset that has a total of 5358 audio tracks and has matched the notes and lyrics to their corresponding timestamps. The extracted lyrics and titles are inputs and critical information for determining the time signature. However, because the DALI dataset lacks time signature information, we performed web scraping to retrieve the information.

### B. Lyrics Patterning

Lyrics patterning contains phrase determination, text preprocessing, keyword extraction, phrase vector patterning, and feature generation. This step is critical to connecting lyrics and rhythm.

*1) Phrase Determination and Text Preprocessing:* To retrieve the keyword vector pattern for each set of lyrics, the lyrics are first split into phrases based on the punctuation and spacing, and the words in each phrase are subsequently tokenized. Stopwords [18], a list of insignificant words, are removed from the text to find the keywords, although we customize our stopwords list since typical text keywords vary from lyrical keywords. For instance, personal subject pronouns such as "I", "we", "you", etc. are usually considered stopwords and thus removed from the text; however, these pronouns are often emphasized in songs, so they should be kept in the text instead due to the value placed on them.

*2) Keyword Extraction:* After the preprocessing, all words are split into syllables and their corresponding stress patterns for each phrase. Stress patterns consist of the stressed and unstressed syllable(s), with each stressed syllable denoted as "1", each stressed syllable with less emphasis denoted as "2", and each unstressed syllable denoted as "0". Due to the Carnegie Mellon University (CMU) pronouncing dictionary [19] used, contractions such as "we've" are split into "we" and "ve", which leads to the code output giving two syllables instead of one. As a result, a customized word list of lingering remnants of verbs such as "ve" from "have" is produced and it is ensured that these remnants are removed. This drastically increases our general algorithm accuracy.

*3) Phrase Vector Patterning:* Using the previously identified phrases, the keywords are categorized by phrase to generate initial vector patterns of stressed syllables of keywords (denoted as phrase vector patterns) before processing them to eliminate noisy data. First, non-keywords are decreased in importance depending on their situation through methods, such as turning each part of their stress patterns to "0". Secondly, all of the stressed and unstressed syllables are separated within each stress pattern so that each syllable is independent within a phrase. These two preprocessing steps help prepare the data for the time signature determination algorithms. Figure 2 shows an example [1] of such phrase vector patterns that are converted from the original lyrics [20].

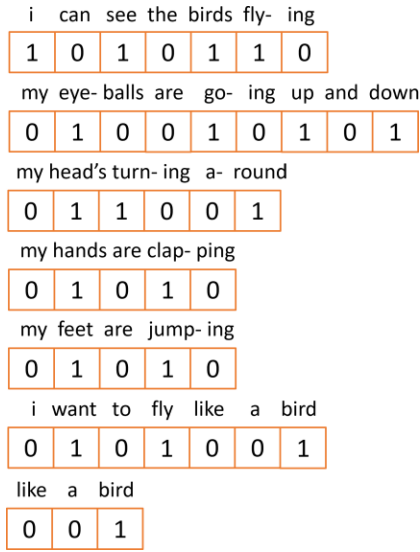

Fig. 2: The example of phrase vector patterns

After this, the vector patterns are optimized with many techniques, taking into account the differences, similarities, and associations between lyrical and musical characteristics. For example, "1" is turned into a "0" if the syllable before it is a "1" (stressed) for some cases. The number of occurrences are totaled for each type of some certain patterns, such as ["0", "0"], ["1", "0"], and ["0", "1", "0"], to find the most frequent patterns. The repetitiveness in the lyrics is also contained in the patterns.

Moreover, we have discovered specific patterns according to time signature conventions that are defined as *stress beat patterns*. For example, in 3/4 time, the pattern is ["1", "0", "0"], and in 4/4 time, the pattern is ["1", "0", "2", "0"], with the first downbeat having a stronger accent than the second downbeat. However, instead of considering that pattern, both ["1", "0"] and ["1", "0", "0", "0"] are utilized as the patterns that represent and constitute the 4/4 time since they are segments of the 4/4 time stress beat pattern and thus easier to find. For simplicity, the patterns such as ["1", "0"], ["1", "0", "0"], and ["1", "0", "0", "0"] are denoted as "10", "100", and "1000".

[1]The lyrics from the song "Birds Are Flying" by Ellie L. Zhang.

**Algorithm 1:** Lyrics Patterning

**Input:** One song's lyrics text $S$.
**Output:** One list of $L$.
1: Initialize $L$ as a list of lyrics vector patterns
2: Convert $S$ into a list of phrases $P$
3: **for** each phrase in $P$ **do**
4:     PhraseVec = GetKeywordVectorPattern(phrase)
5:     OptPhraseVec = Optimize(PhraseVec)
6:     Insert OptPhraseVec into $L$
7: **end for**
8: **return** $L$

Algorithm 1 gives an overview of the method of generating lyrical patterns. For one song, this algorithm is intended to produce a list of lyrical patterns through the input of the lyrics text of the indicated song. Each phrase is converted into a keyword vector pattern by splitting up the stress patterns of all the words in a phrase and labeling non-keywords with "0" to decrease them in importance depending on their situations. The optimization for a song's keyword vector pattern is performed based on both musical and lyrical characteristics. The latent information of the lyrical and musical repetitions is included in the lyrical patterns as well.

*4) Feature Generation:* When the lyrical pattern is generated, a variety of features are generated based on main categories: general and repetitive vector patterns as well as stress beat patterns. The repetitive vector patterns differ from the general vector patterns due to the addition of finding duplicate phrase vector patterns in a song. The data further undergoes statistical calculations, such as count, mean, and mode, which are all based on the phrase vector information. In addition to these patterns, noisy patterns are removed from the data. In the end, the combinations of the types are generated as features. For example, if the general and repetitive vector patterns, "10" and "100" stress beat patterns, and the statistical categories count, mean, and mode are chosen, then the total number of features is 12.

Features that are generated using these methods are able to reveal strong correlations and aid us in deciding on the most critical and influential factors for determining the time signature of a song, especially with the phrase vector pattern information embedded within each feature.

### C. Time Signature Determination

After the features are generated, all features or partial features are selected for machine learning models to determine time signatures.

*1) Treating Imbalanced Data:* Despite having features, the data is largely imbalanced; thus, downsampling, oversampling, or a combination of both techniques are used to resolve the issue. We experimented with multiple different models that handle imbalanced data, and the most prominent oversampling model is the Synthetic Minority Over-sampling Technique (SMOTE) [21], while the most effective undersampling model is Tomek Links [22]. In the end, we utilize a combination of SMOTE and Tomek [23].

SMOTE [21] is an oversampling technique that generates synthetic samples from the minority class. The technique is performed by mainly calculating the distance between a randomly selected minority class data point and its $k$ nearest neighbors for some integer $k$, which results in a random synthetic sample that is between those two points in the feature space. A synthetic sample is created as follows:

$$x_{\text{new}} = x_i + \lambda \times (x_{nn} - x_i)$$

where $x_{new}$ represents the new synthetic sample; $x_i$ represents a single data point from the minority class and is the feature vector that is under consideration; $x_{nn}$ is a randomly selected neighbor from the minority class sample $x_i$ that is typically defined by a K-Nearest-Neighbors (KNN) algorithm; and $\lambda$ represents a random value between 0 and 1 that ensures diversity for the new location of the synthetic sample in the feature space.

Tomek Links [22] is a refined Condensed Nearest Neighbor (CNN) technique [24], as CNN is a systematic majority undersampling technique that aims to produce a subset with the ability to classify the data points from the original dataset correctly. Tomek Links builds onto CNN by removing majority class samples that have the shortest Euclidean distance from the minority class instead of randomly selecting samples with the KNN technique.

As mentioned previously, we ultimately utilize SMOTE-Tomek, a method combining the ability of the SMOTE technique to produce synthetic samples for the minority class data and the ability of the Tomek Links model to remove samples from the majority class that are closest to the minority class data [23]. SMOTE is implemented first to reach the desired sample proportion. After that, it is followed by the implementation of Tomek Links.

Figure 3 is an example of pre-treatment vs. post-treatment of the imbalanced data. Before the data are balanced, the 3/4 time data has 179 counts, which constitutes 4.04% of the total data, and the 4/4 time data has 4249 counts. After applying SMOTETomek, the 3/4 time and 4/4 time data have 4253 and 4249 counts, respectively, with their difference being 4 counts, which demonstrates the effectiveness of the technique.

*2) Building Machine Learning Models:* Multiple classification machine learning models are built based on the selected features to determine time signatures. In this paper, four models are applied: logistic regression [25] [26], decision tree [27] [30], random forest [28] [29], and eXtreme Gradient Boosting (XGBoost) [31].

Logistic regression [25] finds the probability of an event's success and failure. The logistic function is critical to logistic regression. For the decision tree [27] and random forest [28] classifiers, the decision tree classifier utilizes all of the available features of the entire dataset, whereas the random forest classifier randomly selects samples and features to construct decision trees.

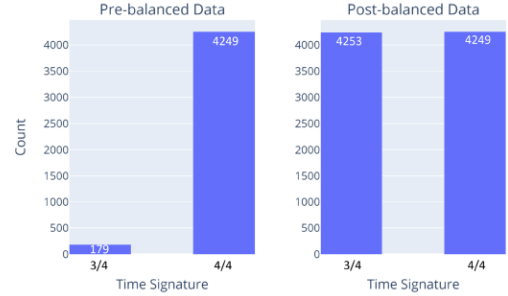

Fig. 3: Effect of Treating Imbalanced Data

XGBoost [31] is an open-source library that effectively implements the Gradient-Boosted Decision Tree (GBDT) algorithms [32] [33]. GBDT can reduce the chances of underfitting and having bias. The gradient-boosted algorithm expands on boosting by utilizing a gradient-descent algorithm over an objective function (used to solve optimization). A regularized objective function is often used to train the ensemble learning method, which helps prevent overfitting. Below is an equation representing the objective function:

$$Objective(\phi) = \sum_{i=1}^{n} l(\hat{y}_i, y_i) + \sum_{k=1}^{K} \Omega(f_k)$$

where $l(\hat{y}_i, y_i)$ represents the loss function which measures the difference between the predicted value $\hat{y}_i$ and the actual value $y_i$, and $\Omega(f_k)$ represents the regularization for the $k$th weak learner.

IV. EXPERIMENTAL RESULTS AND ANALYSIS

The DALI dataset [16] [17] is utilized for this research. It is a multimodal dataset that has a total of 5358 audio tracks with tracked notes and lyrics that have recorded timestamps. However, it does not contain any time signature or key signature information, nor does it have the score information of the songs. We performed automatic web scraping to retrieve the time signature data. 930 songs were removed because 1) some songs contained a time signature other than 3/4 and 4/4 time and/or 2) the time signature simply could not be found. In the end, 4428 songs that are either in 3/4 time or 4/4 time, the most common signatures in lyrical songs, are used for this research. Other time signatures are not included because they have very little data in comparison to the two time signatures. Additionally, all 3/4 time signatures are manually checked by comparing the time signature to the original song to ensure the data's quality.

After the information was retrieved, we discovered that the data was extremely unbalanced, with 4/4 being the majority of the time signatures. Nonetheless, patterns and trends can still be discovered and applied to different time signatures, especially since many time signatures share the same meters and 3/4 and 4/4 times are triple and quadruple meters, respectively.

*1) Exploratory Data Analysis:* We conducted the Exploratory Data Analysis (EDA) to find the insights inside the data.

Figure 4 (a) shows that the 3/4 time and 4/4 time have a significant difference in the number of repetitive pattern counts since the 4/4 time has several thousand more counts than the 3/4 time. However, this is mostly due to the large imbalance in our data. Figure 4 (b) displays the relations between the overall mean of "10" stress beat patterns per phrase and the total average count of the number of phrases with repetitive "10" stress beat patterns. The repetitive pattern count is averaged to reveal a closer comparison between the two time signatures. The "10" occurs more frequently in 4/4 time than 3/4 time since the 4/4 time tends to have 2 subdivisions within the four-beat measure. In 3/4 time, triplets are more common, so a three-beat measure is less commonly divided into groups of 2 notes with equal rhythms. This is demonstrated in the figure, as the "10" pattern has more repeats in 4/4 time than the 3/4 time.

Figure 5 (a) demonstrates the counts of the repetitive patterns based on the corresponding overall mean count for the "100" pattern. Similar to Figure 4 (a), the data is largely imbalanced. In Figure 5 (b), however, the figure displays the relation between the overall mean of "100" stress beat patterns per phrase and the average of the counts of phrases with repetitive "100" stress beat patterns. Contrary to the graph with "10" stress beat patterns, several notable means have more repetitive "100" stress beat patterns, especially for 3/4 time when compared to 4/4 time. This is due to the 3/4 time having a downbeat and two subsequent weak beats, resulting in the similar "100" pattern expressed by the vector patterns that consist of keywords and non-keywords.

Figure 6 consists of two correlation matrices. Figure 6 (a) displays the bi-variate relationships between combinations of variables for overall and repetitive pattern features with "100" stress beat pattern. Except for the count-to-count pair, all other pairs show low correlations. Figure 6 (b) demonstrates the correlation between various statistical types. The main statistics are count and central tendencies mean and mode. The majority of the features have a very low correlation with each other, with many having close to 0 correlation. The highest correlation is between the repetitive mean of the "1000" stress beat pattern and the repetitive count of the "1000" stress beat pattern with a relatively high correlation of 0.76. The two statistical types mean and count have greater correlations than other statistical types, as the highest three correlations are all from these categories. The "1000"-to-"1000" is the highest followed by "100"-"100" and "10"-"10", which both have 0.66 and 0.61 correlations, respectively.

*2) Model Evaluation:* Several model evaluation metrics and 5-fold cross-validations are used in our experiments. Figures 7,

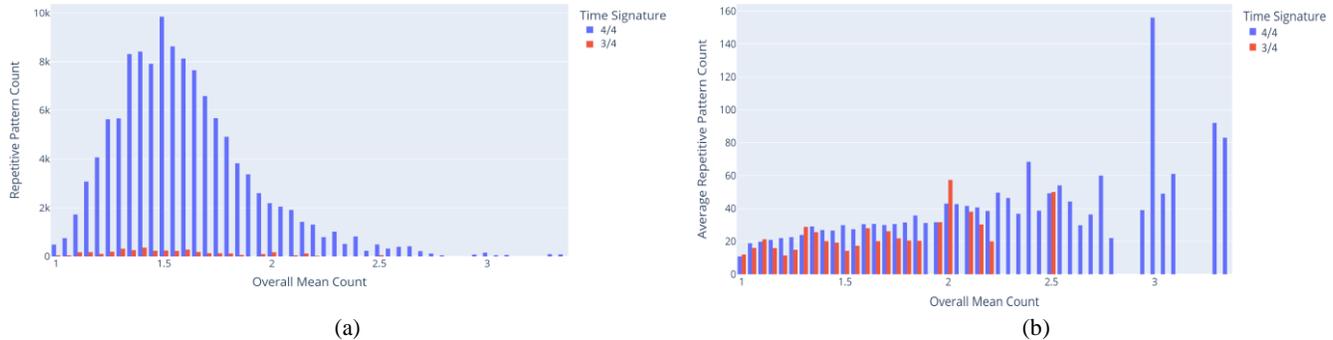

Fig. 4: Histogram of count and average count based on feature categories: repeat, mean, and 10

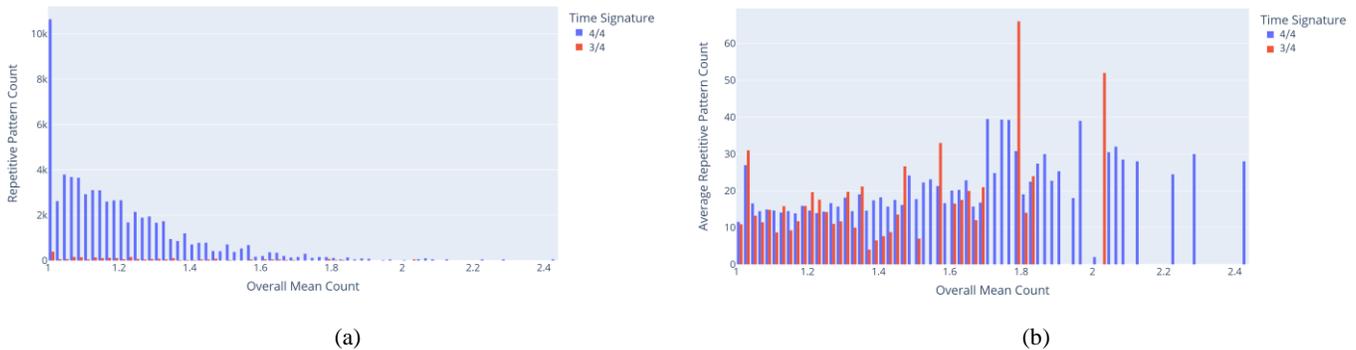

Fig. 5: Histogram of count and average count based on feature categories: repeat, mean, and 100

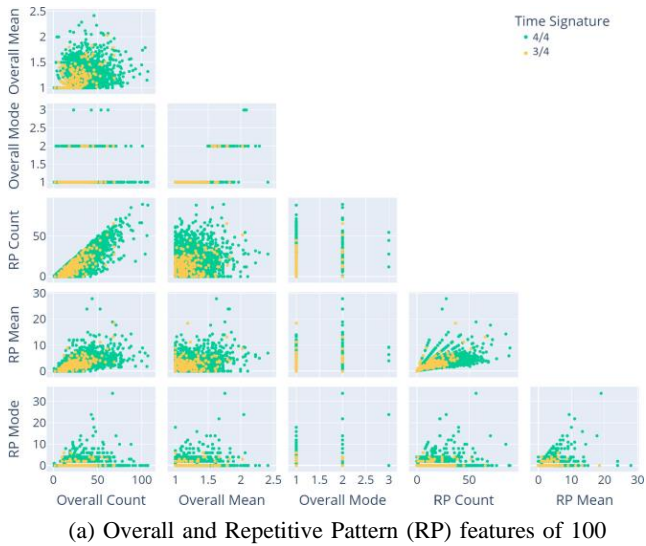

(a) Overall and Repetitive Pattern (RP) features of 100

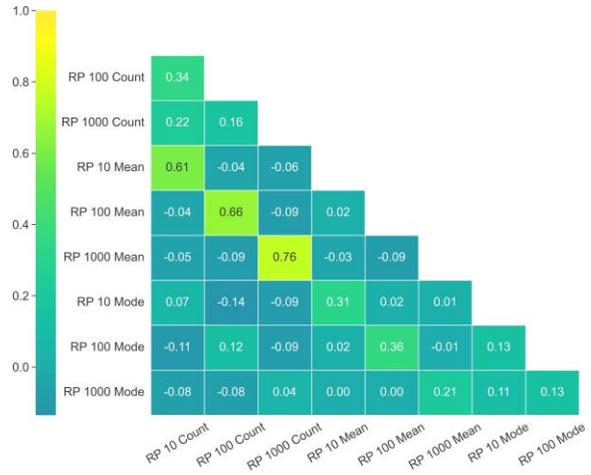

(b) Repetitive pattern features

Fig. 6: Correlation matrices

8, and 9 as well as Table I show the various kinds of metrics are applied when input features containing the combinations of all three categories are selected and some noisy phrase vector patterns are removed. Figure 7 represents the evaluations of the four main machine learning models utilized for time signature determination through confusion matrices. The figure indicates that XGBoost and the random forest classifier have the best fittings since they contain notable correct classifications. In contrast, logistic regression has considerably lower correct classifications.

Figure 8 shows the Receiver Operating Characteristic (ROC) curve and the Area Under the ROC Curve (AUC) scores of the four main machine learning models. The ROC curve is an evaluating metric for classifiers that uses the true positive and false positive rates to plot the curve. The AUC score has a range between 0 and 1, where an AUC score of 1.0 represents a 100% correct prediction, an AUC score of 0.5 represents 50% correction prediction or random guessing, and an AUC score of 0 represents 100% wrong prediction. The figure indicates that XGBoost and the random forest classifier are most suitable for time signature determination since the AUC scores are 0.995 and 0.996, respectively, so the two ROC curves are very close and have a significant amount of overlap in the figure. The AUC score of the decision tree classifier is 0.92. However, the logistic regression model performs poorly in comparison to the others, as the AUC score is only 0.67.

Table I reveals the specific accuracy, precision and recall, F1 score, and the AUC scores for the four classifiers. The F1 score, an evaluation method that utilizes precision and recall to measure the model's accuracy, is very high for the XGBoost, as it is 97.6%. The F1 scores for the decision tree and random forest classifiers are 90.7% and 97.4%, respectively, which both demonstrate high accuracies as well. Logistic regression is the only model with a lower F1 score of 61.7%. Likewise, the actual accuracies are very close to the F1 scores, which indicates that decision tree, random forest, and the XGBoost models are all fitting for time signature determination.

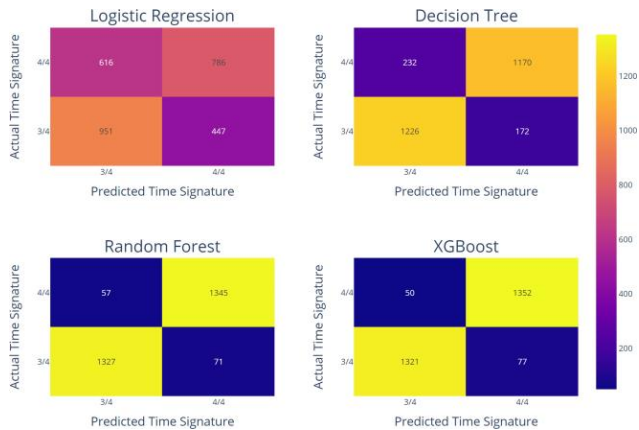

Fig. 7: Confusion matrices of four classifiers

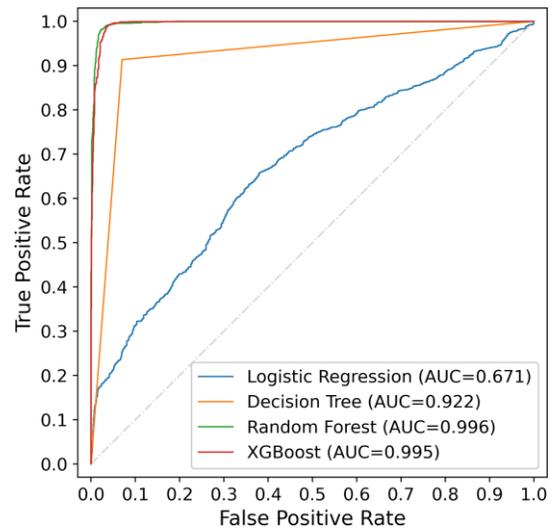

Fig. 8: ROC curves and AUC scores

|                     | Accuracy | Precision | Recall | F1     |
|---------------------|----------|-----------|--------|--------|
| Logistic Regression | 0.6325   | 0.6464    | 0.5897 | 0.6165 |
| Decision Tree       | 0.9082   | 0.9217    | 0.8925 | 0.9068 |
| Random Forest       | 0.9741   | 0.9701    | 0.9786 | 0.9743 |
| XGBoost             | 0.9757   | 0.9666    | 0.9856 | 0.9760 |

TABLE I: The evaluation results of four classifiers

In Figure 9, all of the graphs are based on the generality of the patterns. In particular, (a), (c), and (e) are the cases where all three statistical types and only the specified stress pattern are included. For (b), (d), and (f), they all contain the stress beat patterns but only the specified statistical type. In the graphs with a focus on stress beat patterns, "10" has the highest performance overall, whereas "1000" has the lowest performances overall. For graphs with a focus on the statistical type, the "count" has the highest AUC scores overall, whereas "mode" has the lowest AUC scores overall. Thus, this indicates that the "10" stress beat pattern and the "count" statistical type are the best-performing features. In contrast, the "1000" and the "mode" features have the lowest performances.

*3) Ablation Study for Feature Selection:* Table II shows our ablation study for feature selection. In the table, the stress beat patterns, the statistical types, and the structural types are each listed separately within their category in the table. When the "10", "100", and "1000" patterns are compared separately, "10" and "1000" perform the best and worst, respectively. However, when all three patterns are used, the results are the best. For the statistical types, count performs the best individually, while mode performs the worst. If count and mean are used together, then it gives better results. In terms of the structural types, repetition performs better than general patterns, although if both are used, then they yield the best results. Moreover, when a combination of all of the subcategories are used except for mode, it gives the highest accuracy, precision, F1 score, and AUC score.

Additionally, the highest AUC score displayed in Table II is 0.995, which differs from the 0.996 AUC score shown in Table I. This is because the noisy phrase vector patternsare not removed in this study, whereas some noisy patternsare removed in the previous experiment. Thus, the previous experiment has a slightly higher score.

V. CONCLUSIONS

Musical components such as time signatures continue to be critical to music, yet they are not sufficiently studied enough in AI for music generation, thus hindering researchers' ability to form an algorithmic approach for determining them in new lyrical compositions. Much of the current research in the AI for music field focuses on the broader aspects of music generation rather than the details in structure, rhythm, and melody of which music is comprised. To the best of our knowledge, we are the first to propose robust and explainable methods for automatic time signature generation in new songs by using machine learning and only lyrics, which can serve as one of the critical foundations for AI music generation. Additionally, our efficient method for automatic web scraping has enabled the mass extraction of time signature and key signature information from the Internet, which creates a more well-rounded dataset. We also resolved the significantly imbalanced dataset. The features generated by our novel algorithms for each song encompass lyrical, musical, and statistical information as well. Several machine learning models for time signature determination are trained on the dataset using these novel features as input. Our experimental results reveal 97.6%, 97.4%, and 90.7% F1 scores and 0.995, 0.996, and 0.92 AUC scores for the top three models: XGBoost, the random forest classifier, and the decision tree classifier. Above all, our research delves into an understudied component of musicology and develops a highly accurate time signature determination algorithm when it is only given a set of lyrics, which contributes greatly to the AI for music generation research community. Our future work incorporates time signature determination into more research and applications for AI music, extends our approach to more types of time signatures or meters, as well as inspires and fosters a closer connection between literature and music.

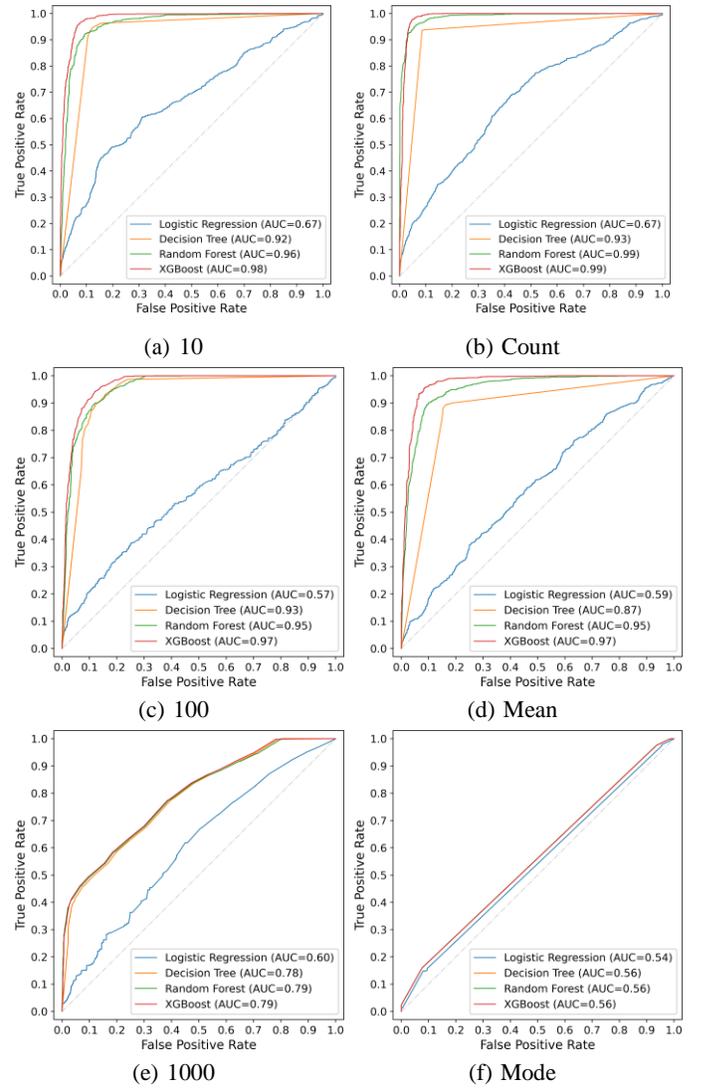

(a) 10

(b) Count

(c) 100

(d) Mean

(e) 1000

(f) Mode

Fig. 9: ROC and AUC scores for the ablation study

| Feature | | | | | | | | Metric | | | | |
|---|---|---|---|---|---|---|---|---|---|---|---|---|
| Overall | Repeat | Count | Mean | Mode | 10 | 100 | 1000 | Accuracy | Precision | Recall | F1 | ROC_AUC |
| * | | * | * | * | * | * | * | 0.9708 | 0.9577 | 0.9859 | 0.9716 | 0.9908 |
| * | | * | * | * | * | * | | 0.9674 | 0.9575 | 0.9796 | 0.9684 | 0.9895 |
| * | | * | * | * | * | | | 0.9352 | 0.9277 | 0.9445 | 0.9360 | 0.9790 |
| * | | * | * | * | | * | | 0.9022 | 0.8814 | 0.9304 | 0.9050 | 0.9673 |
| * | | * | * | * | | | * | 0.6915 | 0.7260 | 0.6163 | 0.6662 | 0.7761 |
| * | | * | | * | * | * | * | 0.9648 | 0.9522 | 0.9796 | 0.9657 | 0.9880 |
| * | | | * | * | * | * | * | 0.9336 | 0.9187 | 0.9550 | 0.9364 | 0.9794 |
| * | | * | | | * | * | * | 0.9655 | 0.9517 | 0.9821 | 0.9666 | 0.9883 |
| * | | | * | | * | * | * | 0.9318 | 0.9144 | 0.9568 | 0.9350 | 0.9759 |
| * | | | | * | * | * | * | 0.5451 | 0.6880 | 0.1647 | 0.2658 | 0.5639 |
| | * | * | * | * | * | * | * | 0.9670 | 0.9571 | 0.9782 | 0.9675 | 0.9899 |
| | * | * | * | * | * | * | | 0.9557 | 0.9458 | 0.9677 | 0.9566 | 0.9870 |
| | * | * | * | * | * | | | 0.9057 | 0.9097 | 0.9017 | 0.9056 | 0.9605 |
| | * | * | * | * | | * | | 0.8703 | 0.8404 | 0.9150 | 0.8760 | 0.9524 |
| | * | * | * | | * | * | * | 0.9654 | 0.9526 | 0.9803 | 0.9663 | 0.9871 |
| | * | | * | * | * | * | * | 0.9154 | 0.9173 | 0.9161 | 0.9166 | 0.970 |
| | * | | | | * | * | * | 0.5576 | 0.5329 | 0.9410 | 0.6804 | 0.5937 |
| | * | * | | * | * | * | * | 0.9612 | 0.9483 | 0.9765 | 0.9621 | 0.9859 |
| | * | | * | * | * | * | | 0.8836 | 0.8840 | 0.8907 | 0.8873 | 0.9495 |
| * | * | * | * | * | * | * | * | 0.9711 | 0.9578 | 0.9859 | 0.9716 | 0.9915 |
| * | * | * | * | * | * | * | | 0.9715 | 0.9576 | 0.9874 | 0.9722 | 0.9899 |
| * | * | * | * | * | * | | | 0.9565 | 0.9502 | 0.9645 | 0.9573 | 0.9842 |
| * | * | * | | * | * | * | | 0.9715 | 0.9576 | 0.9874 | 0.9722 | 0.9899 |
| * | * | * | | * | | * | * | 0.9665 | 0.9611 | 0.9726 | 0.9668 | 0.9926 |
| * | * | * | * | | * | * | * | 0.9741 | 0.9655 | 0.9835 | 0.9744 | 0.9953 |

TABLE II: The ablation study for feature selection. Three colors are used to denote the best values in each metric column: yellow, orange, and blue. Yellow denotes the highest; orange denotes the next few values within the range of the second highest; and blue denotes a few values within the range of the third highest. The asterisks denote the selected types that constitute the feature combination in each row.

## VI. ACKNOWLEDGEMENTS

We would like to acknowledge and thank Ellie L. Zhang for her composition and lyrics that she allowed for this paper.

## REFERENCES

[1] C. C. Liao, D. Liao, and J. Guessford, "Multimodal lyrics- rhythm matching," 2022 IEEE International Conference on Big Data (Big Data), Osaka, Japan, 2022, pp. 3622-3630, doi: 10.1109/BigData55660.2022.10021009.

[2] A. Pikrakis, I. Antonopoulos, and S. Theodoridis, "Music meter and tempo tracking from raw polyphonic audio.", presented at the Proceedings of the 5th International Conference on Music Information Retrieval (ISMIR 2004), Barcelona, Spain, Oct. 2004. doi: 10.5281/zenodo.1416348.

[3] M. Gainza and E. Coyle. "Time signature detection by using a multi resolution audio similarity matrix." Journal of The Audio Engineering Society (2007): n. pag.

[4] M. Gainza, "Automatic musical meter detection," 2009 IEEE International Conference on Acoustics, Speech and Signal Processing, Taipei, Taiwan, 2009, pp. 329-332, doi: 10.1109/ICASSP.2009.4959587.

[5] F. Gouyon and P. Herrera, "Determination of the meter of musical audio signals: seeking recurrences in beat segment descriptors", Presented at the 114th Audio Engineering Society Convention 2003 March 22–25 Amsterdam, The Netherlands.

[6] S. Droit-Volet, D. Ramos, J. L. O. Bueno, and E. Bigand, "Music, emotion, and time perception: the influence of subjective emotional valence and arousal?", Frontiers in Psychology, vol. 4, 2013, doi: https://doi.org/10.3389/fpsyg.2013.00417.

[7] S. Adler, The Study of Orchestration. New York [Etc.] Norton Cop, 2016.

[8] D. Zhang, J. Wang, K. Kosta, J. B. L. Smith, and S. Zhou, "Modeling the rhythm from lyrics for melody generation of pop songs", in Proceedings of the 23rd International Society for Music Information Retrieval Conference, Bengaluru, India, Dec. 2022, pp. 141–148. doi: 10.5281/zenodo.7316616.

[9] D. Naruse, T. Takahata, Y. Mukuta, and T. Harada, "Pop music generation with controllable phrase lengths", in Proceedings of the 23rd International Society for Music Information Retrieval Conference, Bengaluru, India, Dec. 2022, pp. 125–131. doi: 10.5281/zenodo.7316612.

[10] Z. Wang and G. Xia, "MuseBERT: pre-training music representation for music understanding and controllable generation", in Proceedings of the 22nd International Society for Music Information Retrieval Conference, Online, Nov. 2021, pp. 722–729. doi: 10.5281/zenodo.5624387.

[11] Z. Wang, D. Wang, Y. Zhang, and G. Xia, "Learning interpretable representation for controllable polyphonic music generation", in Proceedings of the 21st International Society for Music Information Re-

trieval Conference, Montreal, Canada, Oct. 2020, pp. 662–669. doi: 10.5281/zenodo.4245518.
[12] K. Chen, C. Wang, T. Berg-Kirkpatrick, and S. Dubnov, "Music Sketch-Net: controllable music generation via factorized representations of pitch and rhythm", in Proceedings of the 21st International Society for Music Information Retrieval Conference, Montreal, Canada, Oct. 2020, pp. 77–84. doi: 10.5281/zenodo.4245372.
[13] J. Liu, "Symphony generation with permutation invariant language model", in Proceedings of the 23rd International Society for Music Information Retrieval Conference, Bengaluru, India, Dec. 2022, pp. 551–558. doi: 10.5281/zenodo.7316722.
[14] F. Lerdahl and R. S. Jackendoff, A Generative Theory of Tonal Music, The MIT Press, December 21, 1982. ISBN: 9780262120944.
[15] A. McLeod and M. Steedman, "Meter detection and alignment of MIDI performance", in Proceedings of the 19th International Society for Music Information Retrieval Conference, Paris, France, Sep. 2018, pp. 113–119. doi: 10.5281/zenodo.1492357.
[16] G. Meseguer-Brocal, A. Cohen-Hadria, and Geoffroy Peeters, "DALI: a large dataset of synchronized audio, lyrics and notes, automatically created using teacher-student machine learning paradigm.", in Proceedings of the 19th International Society for Music Information Retrieval Conference, Paris, France, Sep. 2018, pp. 431–437. doi: 10.5281/zenodo.1492443.
[17] G. Meseguer-Brocal, "The DALI dataset". Zenodo, Feb. 26, 2019. doi: 10.5281/zenodo.2577915.
[18] C. Fox, "A stop list for general text", ACM SIGIR Forum. 24 (1–2): 19–21, ISSN 0163-5840, 1989.
[19] K. Lenzo, "The CMU pronouncing dictionary", Carnegie Mellon University. http://www.speech.cs.cmu.edu/cgi-bin/cmudict
[20] E. L. Zhang, "Birds are Flying", Journal of Children's Music, vol. 365, pp. 9, November 2016.
[21] N. V. Chawla, K. W. Bowyer, L. O. Hall, and W. P. Kegelmeyer, "SMOTE: synthetic minority over-sampling technique", Journal of Artificial Intelligence Research, 2002, Vol. 16, pp.321–357. https://doi.org/10.1613/jair.953.
[22] I. Tomek, "Two modifications of CNN," in IEEE Transactions on Systems, Man, and Cybernetics, vol. SMC-6, no. 11, pp. 769-772, Nov. 1976, doi: 10.1109/TSMC.1976.4309452.
[23] G. E. A. P. A. Batista, A. L. C. Bazzan, and M. A. Monard, "Balancing training data for automated annotation of keywords: a case study", Proceedings of the Second Brazilian Workshop on Bioinformatics, 2003, pp. 35–43.
[24] P. E. Hart, "The condensed nearest neighbor rule", IEEE Transactions on Information Theory, 18: 515–516. doi:10.1109/TIT.1968.1054155.
[25] J. Berkson, "Application of the logistic function to bio-assay", Journal of the American Statistical Association, 1944, 39 (227): 357–365.
[26] J. S. Cramer, "The origins of logistic regression", 2002, Vol. 119. Tinbergen Institute. pp. 167–178. doi:10.2139/ssrn.360300.
[27] R. Messenger and L. Mandell, "A modal search technique for predictive nominal scale multivariate analysis", Journal of the American Statistical Association, 1972, vol. 67, pp.768–772.
[28] T. K. Ho, "Random decision forests", Proceedings of the 3rd International Conference on Document Analysis and Recognition, Montreal, QC, Canada, August 1995, pp. 278–282.
[29] T. K. Ho, "The random subspace method for constructing decision forests", IEEE Transactions on Pattern Analysis and Machine Intelligence, 1998, vol. 20 (8), pp. 832–844. doi:10.1109/34.709601.
[30] W. Y. Loh, "Fifty years of classification and regression trees", International Statistical Review, 2014, vol. 82 (3), pp.329-348.
[31] T. Chen and C. Guestrin, "XGBoost: a scalable tree boosting system", KDD '16: Proceedings of the 22nd ACM SIGKDD International Conference on Knowledge Discovery and Data Mining, August 2016, pp.785–794. doi.org/10.1145/2939672.2939785.
[32] J. H. Friedman, "Greedy function approximation: a gradient boosting machine", The Annals of Statistics, 2001, vol. 29 (5), pp. 1189–1232.
[33] J. H. Friedman "Stochastic gradient boosting", Computational Statistics and Data Analysis, 2002, vol. 38 (4), pp. 367–378.